\documentclass[a4paper]{article}

\usepackage{INTERSPEECH2020}
\usepackage{amsmath,graphicx,epstopdf,booktabs,multicol,multirow,enumitem,footnote,type1cm}

\title{Speech-XLNet: Unsupervised Acoustic Model Pretraining \\ For Self-Attention Networks}
\name{\begin{tabular}{c}Xingchen Song$^{1,\#}$,~Guangsen Wang$^{4,\#}$\thanks{\# Work done when the author was in Tencent AI Lab. * Corresponding author.},~Yiheng Huang$^3$,~Zhiyong Wu$^{1,2,*}$,~Dan Su$^3$,~Helen Meng$^{1,2}$\end{tabular}}
\address{
    $^1$Tsinghua Shenzhen International Graduate School, Tsinghua University, Shenzhen, China\\
    $^2$The Chinese University of Hong Kong, Shatin, N.T., Hong Kong SAR, China\\
	$^3$Tencent AI Lab, Shenzhen, China~~$^4$Salesforce Research Asia, Singapore}
\email{sxc19@mails.tsinghua.edu.cn,~~\{zywu,hmmeng\}@se.cuhk.edu.hk,\\    guangsen.wang@salesforce.com,~~\{arnoldhuang,dansu\}@tencent.com}

\begin{document}

\maketitle
\begin{abstract}
Self-attention network (SAN) can benefit significantly from
the bi-directional representation learning through unsupervised pretraining paradigms such as BERT and XLNet. In this paper, we present an XLNet-like pretraining scheme ``Speech-XLNet'' to learn speech representations with self-attention networks (SANs). 
Firstly, we find that by shuffling the speech frame orders, Speech-XLNet serves as a strong regularizer which encourages the SAN network to make inferences by focusing on global structures through its attention weights. Secondly, Speech-XLNet also allows the model to explore bi-directional context information while maintaining the autoregressive training manner. Visualization results show that our approach can generalize better with more flattened and widely distributed optimas compared to the conventional approach.
Experimental results on TIMIT demonstrate that Speech-XLNet greatly improves hybrid SAN/HMM in terms of both convergence speed and recognition accuracy.
Our best systems achieve a relative improvement of 15.2\%
on the TIMIT task. Besides, we also apply our pretrained model to an End-to-End SAN with WSJ dataset and WER is reduced by up to 68\% when only a few hours of transcribed data is used.
\end{abstract}

\section{Introduction}

Recently, unsupervised representation learning paradigms such as BERT~\cite{devlin2019bert}, XLNet~\cite{DBLP:conf/nips/YangDYCSL19} have been highly successful for language modeling with self-attention network (SAN) such as transformer~\cite{vaswani2017attention}. More specifically, SAN can benefit from the bi-directional context representation learning through unsupervised pretraining with large-scale unlabeled data before finetuning with labeled data. For automatic speech recognition (ASR), SANs have been introduced for acoustic modeling in either attention framework (also known as speech transformer~\cite{Dong2018SpeechTransformerAN}) or with CTC loss~\cite{graves2006connectionist,Salazar2019SelfattentionNF}. 

Unsupervised speech representation learning has also been investigated by learning to predict the future either in latent space~\cite{Schneider2019,Baevski2020vq-wav2vec} or real frames~\cite{Chung2019,Mockingjay,Lian2019,ling2019deep,wang2020unsupervised}. The pretrained networks are then used as feature extractors ahead of various downstream tasks,~
such as speaker and phone recognition~\cite{Chung2019,Mockingjay}, emotion recognition~\cite{Lian2019} and speech recognition~\cite{Schneider2019,Baevski2020vq-wav2vec,ling2019deep,wang2020unsupervised}.
Among those works, Autoregressive Predictive Coding (APC) was first employed in~\cite{Chung2019} for speech recognition, where an RNN-based model  was pretrained in an autoregressive (AR) manner to encode temporal information of past acoustic sequence. Wav2vec~\cite{Schneider2019} proposed a
multi-layer convolutional neural network optimized via a noise
contrastive binary classification, which is also uni-directional. The constraints on AR models are two-fold:~1) Because neighboring speech frames are highly correlated, exploiting such local smoothness might be already enough to predict next frames. Therefore, without regularization, the pretraining process may have difficulties in capturing global structures.~2) AR pretraining also suffers from the lack of bi-directional context information which is helpful in sequence-to-sequence tasks.
Unlike left-to-right uni-directional approaches that only consider past sequences, The works of~\cite{Baevski2020vq-wav2vec,Mockingjay,wang2020unsupervised} proposed a BERT-like autoencoding (AE) scheme to train a bi-directional speech representation model. However, the zero-masked frames used by BERT during pretraining are absent from real data in the finetuning process, thus lead to a pretrain-finetune discrepancy~\cite{DBLP:conf/nips/YangDYCSL19}. Moreover, BERT assumes that the predicted frames are independent of each other given the unmasked frames, so it is unable to model the joint probability distributions using the product rule as used in the AR models~\cite{DBLP:conf/nips/YangDYCSL19}.

Given the success of XLNet~\cite{DBLP:conf/nips/YangDYCSL19} in language modeling, in this paper, we present ``Speech-XLNet'', a permutation-based AR acoustic model pretraining scheme for SANs.  Instead of using a fixed forward order as in AR models~\cite{Schneider2019,Chung2019,Lian2019}, Speech-XLNet maximizes the expected log likelihood of a speech feature sequence with respect to all possible permutations of the factorization order. We conjecture that by shuffling the frame orders, the permutation helps to capture bi-directional contextual information and global structures, leveraging the best of both AE acoustic modeling and AR while avoiding their drawbacks. 

We evaluated the proposed method on two systems. For hybrid SAN/HMM, we reduce phone error rate (PER) from 15.1\% to 12.8\% on TIMIT-test. For end-to-end (E2E) model, we obtain relative word error rate reduction (WERR) from 68.3\% to 8.5\% on WSJ-eval92 when the amount of labeled WSJ-si284 increased from 7h to 81h. Besides, we also provide some visualizations to demonstrate the generalization properties of our proposed method.

\section{Related work}

The most related work to this paper is XLNet~\cite{DBLP:conf/nips/YangDYCSL19}. Our scheme is different from XLNet mainly in three aspects.~1) We don't use the next sentence prediction and segment recurrence mechanism as mentioned in~\cite{DBLP:conf/nips/YangDYCSL19}, both methods are used to extract long-distance dependencies across consecutive sentences. However, in ASR datasets, different sentences are generally independent of each other and have no contextual relationships.~2) We mask and reconstruct speech frames rather than word tokens. In other words, we adapt XLNet from a classification task to a regression task.~3) For more robust pretraining, instead of performing permutation once during data preprocessing in original XLNet,  we adopt a \textit{\textbf{dynamic}} permutation strategy where we regenerate a permutation order every time a speech sequence is fed to the trainer. Thus, we have effectively increased the amount of training data by using different permutation for every sequence fed into each training epoch. 



More recently, capturing bi-directional context information for AR acoustic models has been explored in \cite{ling2019deep}. The authors proposed an LSTM-based pretraining scheme, in which the outputs of two LSTMs (one left-to-right and one right-to-left) were concatenated to reconstruct the selected frames.
Because the two uni-directional LSTMs were \textit{\textbf{independently}} trained, forward and backward contexts cannot see each other across different layers. The concatenated output thus can be seen as a \textit{\textbf{shallow bi-directional}} representation. On the other hand, in ``Speech-XLNet'',  the forward and backward contexts are \textit{\textbf{jointly}} learned within a single model. Since left and right content have access to each other, they are more integrated, yielding a \textit{\textbf{deeper bi-directional}} representation.

\section{Speech-XLNet}
\begin{figure}[htbp]
	\begin{minipage}[b]{1.\linewidth}
		\centerline{\includegraphics[width=8cm]{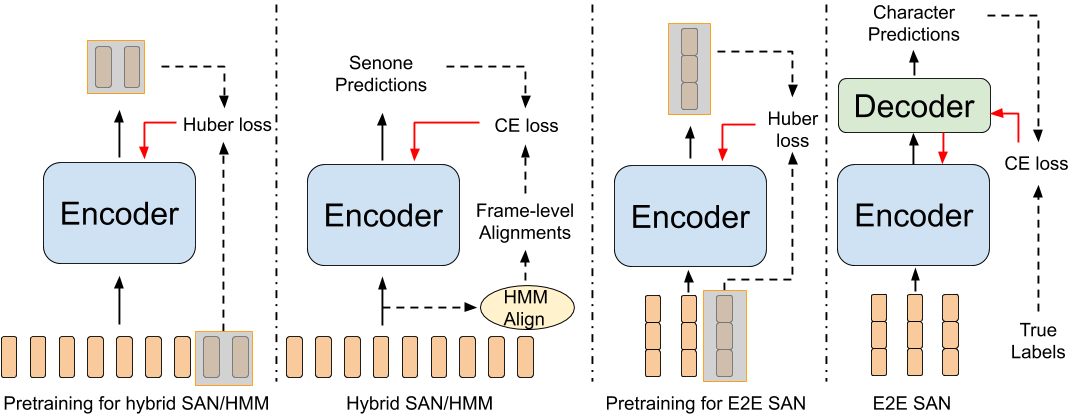}}
	\end{minipage}
	\caption{Speech-XLNet system overview. The red arrow indicates the back propagation path of the gradients.}
	\label{fig:model-framework}
\end{figure}

The system overview of the proposed Speech-XLNet is shown in Fig.\ref{fig:model-framework}. The architectures of encoder and decoder are the same as XLNet~\cite{DBLP:conf/nips/YangDYCSL19} and Transformer~\cite{vaswani2017attention}, respectively. The main difference between the hybrid SAN/HMM and the end-to-end SAN in the pretraining process is that the hybrid model does not perform frame-stacking\cite{8682586} on input features, this is because we need to obtain frame-level alignments for downstream ASR task in hybrid SAN/HMM.

\subsection{Pretrain objective function}

Different from the density estimation process in XLNet, Speech-XLNet aims to ``predict'' the next acoustic frame. Specifically,
let $\mathcal{O}_T$ be the set of all possible permutations of a length-$T$ frame sequence.  Let $\mathbf{O}=[\mathbf{o}_1,\mathbf{o}_2,...,\mathbf{o}_T]$,  $\mathbf{o}_t$ and $\mathbf{o}_{<t}$  denote the $t$-th element and the first $t-1$ elements of a permutation $\mathbf{O}\in \mathcal{O}_T$. The permutation acoustic modeling objective can be expressed as:
\begin{equation}
\min_{\bm{\theta}} \mathbb{E}_{\mathbf{o} \sim \mathcal{O}_T}\bigg[\sum_{t=1}^T \mathcal{L}({\mathbf{o}_t}, \hat{\mathbf{y}}_{t|\mathbf{o}_{<t}})\bigg]
\end{equation}
where $\bm{\theta}$ denotes parameter set, $\hat{\mathbf{y}}_{t|\mathbf{o}<t}$ is the predicted frame given the previous frames of the permutation order $\mathbf{o}_{<t}$. 
In our preliminary experiments, we found that both L1 and L2 loss fail to converge due to permutation, thus we adopted the smooth Mean Absolute Error (MAE) loss (Huber loss) as in~\cite{Ren2015FasterRT} as our pretraining loss, which is a combination of L1 and L2 losses as given below:
\begin{equation*}
     \begin{gathered}
    \mathcal{L}({\mathbf{o}_t}, \hat{\mathbf{y}}_{t|\mathbf{o}_{<t}}) = 
        \begin{cases}
        \frac{1}{2\delta}({\mathbf{o}_t}- \hat{\mathbf{y}}_{t|\mathbf{o}_{<t}})^2,&  |{\mathbf{o}_t}- \hat{\mathbf{y}}_{t|\mathbf{o}_{<t}}| < \delta \\
        |{\mathbf{o}_t}- \hat{\mathbf{y}}_{t|\mathbf{o}_{<t}}| -  \frac{\delta}{2},& \text{Otherwise }
        \end{cases}
    \end{gathered}
\end{equation*}
where $\delta$ is a scalar to balance the L1 and L2 losses.
As $\bm{\theta}$ is shared across all orders in $\mathcal{O}_T$, $\mathbf{o}_t$ can see all other frames $\mathbf{o}_{t'} \ne \mathbf{o}_t$ via permutations, which encourages the model to learn bi-directional contexts and longer span structures. 

To reduce the optimization difficulty, we use similar setup to~\cite{DBLP:conf/nips/YangDYCSL19}, where only the last 20\% of frames from the tail portion of the permuted sequence are chosen to be predicted. Formally, let $T$ be the sequence length and $e$ be the number of selected frames. 
The objective function thus becomes:
\begin{equation}\label{eq:tail_pred}
\min_{\bm{\theta}} \mathbb{E}_{\mathbf{o} \sim \mathcal{O}_T}\bigg[\sum_{t=T-e+1}^T \mathcal{L}({\mathbf{o}_t}, \hat{\mathbf{y}}_{t|\mathbf{o}<t})\bigg]
\end{equation}

\subsection{Two-stream self-attention}

Same as XLNet, the permutation is achieved via attention masks while keeping the original sequence order. It introduces the ambiguity in target prediction since the standard hidden representation does not have the information of which position it will predict. This can be addressed by the target-aware representations via a two-stream attention mechanism~\cite{DBLP:conf/nips/YangDYCSL19}, namely the content stream and the query stream. The two streams of representations are schematically updated with a shared set of parameters (Note that the query stream is only used in pretraining and discarded during finetuning). Due to space limitations, the details are omitted and interested readers can refer to~\cite{DBLP:conf/nips/YangDYCSL19}.

\section{Experiments and Analyses}

Our model is implemented with PyTorch~\cite{paszke2019pytorch}. All the acoustic features used in this paper are the 40-dimensional log-Mel filterbanks extracted using Kaldi~\cite{povey2011kaldi} with global cepstral mean and variance normalization.

\subsection{Hybrid ASR on TIMIT}

The encoder used for hybrid SAN/HMM consists of a stack of six self-attention blocks, with a per-block configuration of 8 attention heads, the model dimension  $d_{model}=512$ and feed-forward inner-layer dimension $d_{inner}=2048$.
Dropout and Huber loss $\delta$ is set to 0.1 and 1.0, respectively. The network is optimized using Adam~\cite{DBLP:journals/corr/KingmaB14} with $\beta_1=0.9$, $\beta_2=0.999$, $\epsilon=1e-6$ and a weight decay of 0.01. No delta or frame stacking is used.

For pretraining, we use a pool of Librispeech~\cite{Panayotov2015LibrispeechAA}, TED-LIUM release2~\cite{Rousseau2012TEDLIUMAA} and the WSJ-si284 corpora. The network was trained with a total of 1M steps with 115k warm-up steps and a linear learning rate decay~\cite{devlin2019bert}, which translates to a peak learning rate of $6e-4$. The pretraining was conducted with a batch size of 6000 frames and the model parameters were updated with a gradient accumulation of 10 batches. 

The pretrained SAN was then finetuned with Cross Entropy (CE) loss under the hybrid SAN/HMM setup with a batch size of 4000 frames. A total of 10k steps(nearly 40 epochs) were conducted with 1k warm-up steps. Same hyper-parameters were used for training the baseline SAN with randomly initialized weights.

The frame-level alignments were obtained using the Kaldi s5 recipe with a triphone GMM/HMM with 1936 senone clusters.  For decoding, a bigram phone language model trained on the training data transcriptions was used. 
\begin{figure}[htbp]
	\begin{minipage}[b]{1.0\linewidth}
		\centerline{\includegraphics[width=7.5cm]{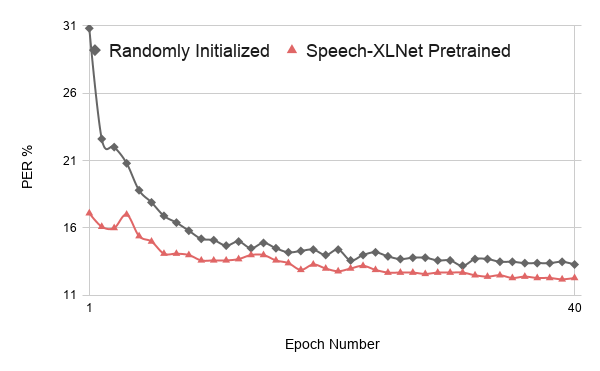}}
	\end{minipage}
	\caption{PERs(\%) on TIMIT dev-set.}
	\label{fig:per}
\end{figure}

The PER trend on the dev set is depicted in Fig.\ref{fig:per}. We can clearly observe that the pretrained SAN converges much faster than the one with randomly initialized weights at early epochs. More importantly, the pretrained SAN also consistently outperforms the baseline system. 
\begin{figure*}[htbp]
	\begin{minipage}[b]{0.48\linewidth}
		\centerline{\includegraphics[scale=.6]{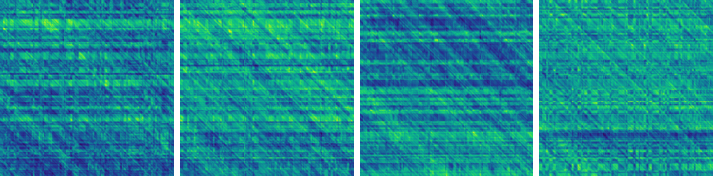}}
		\centerline{(a) Randomly initialized weights, start of training$\footnotemark[2]$}
		\centerline{\includegraphics[scale=.6]{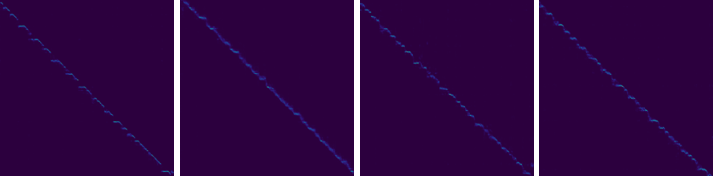}}
		\centerline{(c) Randomly initialized weights, end of training$\footnotemark[2]$}
		
	\end{minipage}
	\hfill
	\begin{minipage}[b]{0.48\linewidth}
		\centerline{\includegraphics[scale=.6]{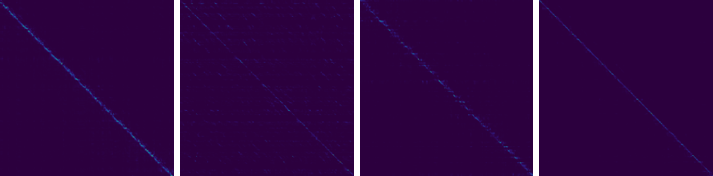}}
		\centerline{(b) Speech-XLNet pretrained weights, start of finetuning$\footnotemark[2]$}
		\centerline{\includegraphics[scale=.6]{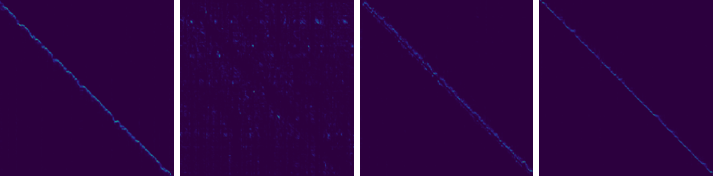}}
		\centerline{(d) Speech-XLNet pretrained weights, end of finetuning$\footnotemark[2]$}
	\end{minipage}
	\caption{Plots of attention scores of the last self-attention block for a TIMIT sentence ``FEDW0-SX364''.}
	\label{fig:att}
\end{figure*}
We also present some PER comparisons with different hyper-parameter setups in Table~\ref{table:lr}. It is well known that SANs are very sensitive to hyper-parameters. Therefore, huge efforts have to be devoted to architecture engineering and hyper-parameter tuning. This can also be seen in Table~\ref{table:lr} as different learning rates affect the PERs of the randomly initialized SANs significantly. On the contrary, for pretrained SANs, the performance is much more stable.
\begin{table} [!htp]
	\caption{PER(\%) comparison of different learning rates.}
	\label{table:lr}
	\begin{minipage}[b]{1.\linewidth}
		\centerline{
			\scalebox{0.9}{
				\begin{tabular}{ccccc}
					\toprule[1.5pt]
					\multirow{2}{*}{learning rate}&
					\multicolumn{2}{c}{Randomly Init}&\multicolumn{2}{c}{ Speech-XLNet Pretrained}\\
					\cmidrule(lr){2-3} \cmidrule(lr){4-5}
					& dev & test & dev & test\\
					\midrule
					3e-5 & 16.7 & 18.4 & 11.8 & 12.9 (29.9\%$\footnotemark[1]$) \\
					1e-4 & 14.7 & 16.8 & 11.8 & {\bf12.8} (23.8\%)\\
					2e-4 & 14.5 & 16.1 & 12.0 & 13.0 (19.3\%)\\
					1e-3 & 13.2 & {\bf 15.1} & 12.0 & 13.1 (13.2\%)\\
					2e-3 & diverge & diverge & 13.1 & 14.1 ($\infty$)\\
					\bottomrule[1.5pt]				
				\end{tabular}
			}
		}
	\end{minipage}
\end{table}
\footnotetext[1]{Numbers inside parentheses are relative improvement over randomly initialized weights with the same training hyper-parameter setup.}

With permutation-based  pretraining, the attention weights are more focused on learning global information.
To validate our conjectures, we visualized the attention scores of the last self-attention block with different configurations in Fig.\ref{fig:att}.
By comparing Fig.\ref{fig:att}~(a) and Fig.\ref{fig:att}~(b), Some prior knowledge of how to represent the data distribution has been clearly learned by Speech-XLNet, which helps the subsequent finetune process converge faster. Furthermore, at the end of training with randomly initialized weights,
the attention scores of all heads manifest a clear diagonal pattern as shown in Fig.\ref{fig:att}~(c) , which means only limited context around the current frame is explored. On the other hand, it is interesting to note in Fig.\ref{fig:att}~(d), the attention scores of the second head are all off-diagonal to a large margin with fairly spread out probability distributions. This may indicate that the attention distribution has learned some ``global'' structures which is useful for the subsequent  classification task.
\begin{table} [!htp]
	\caption{PER comparison with previous pretrain methods. We approximate the number of parameters based on the description in the previous studies.}
	\label{table:compare_sota_timit}
	\begin{minipage}[b]{1.\linewidth}
		\centerline{
			\scalebox{0.7}{
				\begin{tabular}{cccc}
					\toprule[1.2pt]
					Pretrain Method & Pretrain Data & Pretrain Params & Dev/Test PER(\%)\\
					\midrule
					VQ-Wav2vec (\cite{Baevski2020vq-wav2vec}) & libri (960h) & 34M & 15.34 / 17.78\\
					RBM-DBN (\cite{Michalek2018ASO}) & timit (8h) & $\approx$ 34.2M & 15.90 / 16.80 \\
					Ours (Randomly Init) & - & 19.9M & 13.20 / 15.10 \\
					Wav2vec (\cite{Schneider2019}) & libri+wsj (1041h) & 34M & 12.90 / 14.70\\
					Ours (Pretrained) & libri+wsj+ted (1248h) & 19.9M & 11.70 / 12.80\\
					VQ-Wav2vec+BERT (\cite{Baevski2020vq-wav2vec}) & libri (960h) & $\approx$ 71.8M & 9.64 / 11.64\\
					\bottomrule[1.2pt]				
				\end{tabular}
			}
		}
	\end{minipage}	
\end{table}

Lastly, we give a PER comparison with other approaches in Table~\ref{table:compare_sota_timit}. Unsupervised pretraining can be traced back to deep belief network (DBN)~\cite{Hinton2012DeepNN} proposed at the early adoption of DNNs. DBN is a generative model  built with  stacks of restricted Boltzmann machines (RBMs). By comparing the second and third rows, our model outperforms RBM-based method even without pretraining. We also gather the recently published PERs in the last three rows. Our pretraining method achieves the PER
of 12.8\%, which is a comparable performance with state-of-the-art result (Although we use 288 hours more data for pretraining, the encoder parameters used in VQ-Wav2vec+BERT~\cite{Baevski2020vq-wav2vec} are nearly four times than ours).

\subsection{End-to-End transformer ASR on WSJ}

We further adapt Speech-XLNet to the transformer-based system to study the impact of ``Speech-XLNet'' for end-to-end ASR.

The transformer  contains 12 encoder blocks and 6 decoder blocks. Each block has four attention heads and the model dimension $d_{model}$ is 256. $d_{inner}$ is set to 2048 (More details of decoder block can be found in~\cite{vaswani2017attention}). The output alphabet of target text consists of 32 classes, including 26 uppercase letters, apostrophe, period, space, unknown token, sentence start and end tokens.

During both pretraining and finetuning process, three consecutive frames are stacked followed by a frame-skipping rate of three.
Adam \cite{DBLP:journals/corr/KingmaB14} optimizer ($\beta_1=0.9$, $\beta_2=0.98$, $\epsilon=1e-9$) is used to train our model for a total of 420k steps with a batch size of 10000 frames. 
We find that the encoder-decoder architecture cannot converge with a linear learning rate decay thus we use the same schedule as described in~\cite{vaswani2017attention}:
\begin{equation}
lrate = k*d_{model}^{0.5}*min(n^{-0.5}, n*warmup_n^{-1.5})
\end{equation}
where $n$ is the step number. We choose $warmup_n=140000$ for all experiments. We set $dropout=0.1, k=2$ for pretraining and $dropout=0.2, k=11$ for finetuning.  Our baseline model was trained from scratch with the same setups as in the finetuning process. In order to do evaluation, we average the last ten models with the lowest loss on validation set. 
The beam search is carried out with a beam size of 15 and length penalty of $0.6$. Language model was not used during decoding.

To study the impacts of the pretraining with respect to the varying amount of labeled training data available, we train the model with reduced amount of WSJ-si284 and report WERs from 5 of our settings:~1)~the \textit{\textbf{SCRATCH}} where we train the baseline E2E model with randomly initialized weights.~2)~\textit{\textbf{WSJ-Perm}} and~3)~\textit{\textbf{LIBRI-Perm}} where we use 81h WSJ-si284 and 960h Librispeech to pretrain the encoder with ``Speech-XLNet'' and then fintune it with additional decoder.~4)~\textit{\textbf{WSJ-NoPerm}} and~5)~\textit{\textbf{LIBRI-NoPerm}} where we discard permutation during pretraining, which can be roughly seen as uni-directional autoregressive models.

The performance comparison of different methods is shown in Table~\ref{table:compare_sota_wsj}, 
from which 4 conclusions can be deduced.~1) Similar to the hybrid task, finetuning with Speech-XLNet consistently outperforms the network trained from scratch.~2) The smaller amount of labelled data for downstream ASR task, the bigger improvement is achieved.~3) Downstream ASR process can benefit from more unsupervised pretraining data as
\textit{\textbf{LIBRI-Perm}} outperforms \textit{\textbf{WSJ-Perm}}.~4) Permutaion procedure helps to capture bi-directional context information as \textit{\textbf{WSJ-Perm}} outperforms \textit{\textbf{WSJ-NoPerm}} and \textit{\textbf{LIBRI-Perm}} outperforms \textit{\textbf{LIBRI-NoPerm}} for all cases.
More interestingly, with 14h and 81h of labels available, \textit{\textbf{WSJ-Perm}} even  outperforms \textit{\textbf{LIBRI-NoPerm}} that uses 879 hours more data for pretraining, indicating the advantages of the permutation-based pretraining.
\begin{table} [!htp]
	\caption{WER(\%) on eval92 across different amount of si284.}
	\label{table:compare_sota_wsj}
	\begin{minipage}[b]{1.\linewidth}
		\centerline{
			\scalebox{0.75}{
				\begin{tabular}{ccccc}
					\toprule[1.2pt]
					 & 7h & 14h & 30h & 81h\\
					\midrule
					\textit{\textbf{SCRATCH}} & 96.07 & 30.43 & 18.08 & 13.70\\
					\textit{\textbf{WSJ-NoPerm}} & 33.01(65.6\%$\footnotemark[1]$) & 22.4(26.4\%) & 16.16(10.6\%) & 13.11(4.3\%) \\
					\textit{\textbf{WSJ-Perm}} & 31.92(66.8\%) & 22.2(27.0\%) & 15.97(11.7\%) & 12.87(6.1\%)\\
					\textit{\textbf{LIBRI-NoPerm}} & 30.85(67.9\%) & 22.65(25.6\%) & 15.81(12.6\%) & 13.17(3.9\%)\\
					\textit{\textbf{LIBRI-Perm}} & {\bf 30.43}(68.3\%) & {\bf 21.9}(28.0\%) & {\bf 15.7}(13.2\%) & {\bf 12.53}(8.5\%)\\
					\bottomrule[1.2pt]				
				\end{tabular}
			}
		}
	\end{minipage}
\end{table}	

To understand why the model pretrained by Speech-XLNet tends to have better generalization capability on unseen test-set, we further inspect the loss landscapes of the model with or without Speech-XLNet pretraining. Previous researches~\cite{li2018visualizing,hao2019visualizing} show
that the flatness and wideness of a local optimum highly correlate with the generalization capability, i.e., more flattened optima lead to better generalization. This inspires us to visualize the local optima as defined in~\cite{hao2019visualizing}. Formally, let $\theta_0$ denote the initialized parameters. For finetuning, $\theta_0$ represents the pretrained parameters. For training from scratch, $\theta_0$ represents the randomly initialized parameters. After
finetuning, the model parameters are updated to
$\theta_1$. The loss curve aims to
quantify the loss values along the optimization direction (i.e., from $\theta_0$ to $\theta_1$).
The loss curve function $f(\alpha)$ is defined as a linear interpolation of  $\theta_0$ and $\theta_1$:
\begin{equation}
f(\alpha) = \mathcal{J}(\theta_0 + \alpha\cdot\delta)
\end{equation}
where $\alpha$ is a scalar parameter, $\delta = \theta_1-\theta_0$ is the optimization direction, and $\mathcal{J}(\theta)$ is the loss function under the model parameters $\theta$. In our experiments, we set the range of $\alpha$ to [−4, 4] and sample 40 points for each axis. Note that we only consider the pretraining related parameters in $\theta_0$ and $\theta_1$, so $\delta$ only
indicates the updates of the original Speech-XLNet encoder parameters, \emph{i.e.,} the effect of the transformer decoder layers is eliminated by freezing their parameters.

\begin{figure}[htbp]
	\vspace{-12pt}
	\begin{minipage}[b]{1.0\linewidth}
		\centerline{\includegraphics[width=6.0cm]{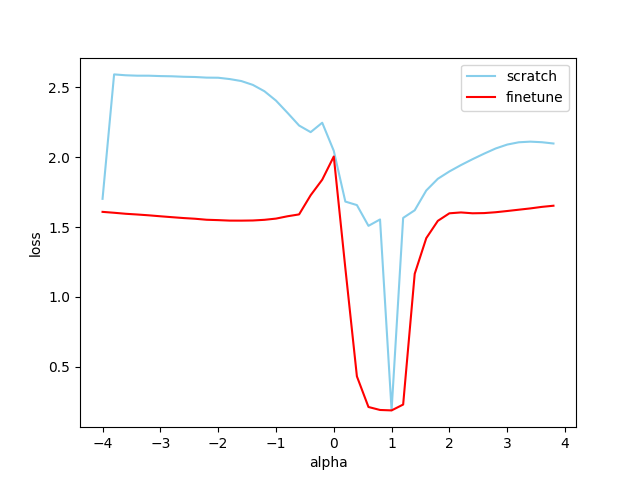}}
	\end{minipage}
	\caption{Loss landscapes on eval92, the optima of training from scratch are sharper than finetuning.}
	\label{fig:curve_loss}
\end{figure}

As shown in Fig.\ref{fig:curve_loss}, the optima of finetuning are more flattened and widely distributed around $\alpha=1$, which indicates Speech-XLNet does help the model to generalize better.
\footnotetext[2]{We omit the remaining four heads as they have the same trend.}


\vspace{-12pt}
\section{Conclusions and Discussions}

In this work, we present Speech-XLNet for unsupervised speech representation learning of self-attention networks (SANs). The effectiveness of our method was evaluated on both hybrid SAN/HMM and transformer-based E2E systems. Experiments demonstrate that with permutation-based pretraining, the SANs tend to focus on more global information and bi-directional context with a faster convergence speed. Furthermore, we show that our pretraining method helps to generalize better with wider and more flattened optimas. 
For futher work, we will evaluate the proposed Speech-XLNet on industry-level large vocabulary continuous speech recognition (LVCSR) with tens of thousands hours of data. Moreover, ``Speech-XLNet'' is also quite appealing for various other scenarios where the amount of labelled data is limited. For example, for code-switching ASR~\cite{Shan2019InvestigatingES}, we can pretrain the model with a pool of large amount of mono-lingual data of all target languages and finetune it with the limited code-switching data.



\bibliographystyle{IEEEtran}

\bibliography{mybib}

\begin{thebibliography}{10}
\providecommand{\url}[1]{#1}
\csname url@samestyle\endcsname
\providecommand{\newblock}{\relax}
\providecommand{\bibinfo}[2]{#2}
\providecommand{\BIBentrySTDinterwordspacing}{\spaceskip=0pt\relax}
\providecommand{\BIBentryALTinterwordstretchfactor}{4}
\providecommand{\BIBentryALTinterwordspacing}{\spaceskip=\fontdimen2\font plus
\BIBentryALTinterwordstretchfactor\fontdimen3\font minus
  \fontdimen4\font\relax}
\providecommand{\BIBforeignlanguage}[2]{{%
\expandafter\ifx\csname l@#1\endcsname\relax
\typeout{** WARNING: IEEEtran.bst: No hyphenation pattern has been}%
\typeout{** loaded for the language `#1'. Using the pattern for}%
\typeout{** the default language instead.}%
\else
\language=\csname l@#1\endcsname
\fi
#2}}
\providecommand{\BIBdecl}{\relax}
\BIBdecl

\bibitem{devlin2019bert}
J.~Devlin, M.-W. Chang, K.~Lee, and K.~Toutanova, ``Bert: Pre-training of deep
  bidirectional transformers for language understanding,'' in \emph{NAACL-HLT},
  2019, pp. 4171--4186.

\bibitem{DBLP:conf/nips/YangDYCSL19}
Z.~Yang, Z.~Dai, Y.~Yang, J.~G. Carbonell, R.~Salakhutdinov, and Q.~V. Le,
  ``Xlnet: Generalized autoregressive pretraining for language understanding,''
  in \emph{NeurIPS}, 2019, pp. 5754--5764.

\bibitem{vaswani2017attention}
A.~Vaswani, N.~Shazeer, N.~Parmar, J.~Uszkoreit, L.~Jones, A.~N. Gomez,
  L.~Kaiser, and I.~Polosukhin, ``Attention is all you need,'' in
  \emph{NeurIPS}, 2017, pp. 5998--6008.

\bibitem{Dong2018SpeechTransformerAN}
L.~Dong, S.~Xu, and B.~Xu, ``Speech-transformer: A no-recurrence
  sequence-to-sequence model for speech recognition,'' in \emph{ICASSP}, 2018,
  pp. 5884--5888.

\bibitem{graves2006connectionist}
A.~Graves, S.~Fern{\'a}ndez, F.~Gomez, and J.~Schmidhuber, ``Connectionist
  temporal classification: labelling unsegmented sequence data with recurrent
  neural networks,'' in \emph{ICML}, 2006, pp. 369--376.

\bibitem{Salazar2019SelfattentionNF}
J.~Salazar, K.~Kirchhoff, and Z.~Huang, ``Self-attention networks for
  connectionist temporal classification in speech recognition,'' in
  \emph{ICASSP}, 2019, pp. 7115--7119.

\bibitem{Schneider2019}
S.~Schneider, A.~Baevski, R.~Collobert, and M.~Auli, ``Wav2vec: Unsupervised
  pre-training for speech recognition,'' in \emph{Interspeech}, 2019, pp.
  3465--3469.

\bibitem{Baevski2020vq-wav2vec}
A.~Baevski, S.~Schneider, and M.~Auli, ``Vq-wav2vec: Self-supervised learning
  of discrete speech representations,'' in \emph{ICLR}, 2020.

\bibitem{Chung2019}
Y.-A. Chung, W.-N. Hsu, H.~Tang, and J.~Glass, ``{An unsupervised
  autoregressive model for speech representation learning},'' in
  \emph{Interspeech}, 2019, pp. 146--150.

\bibitem{Mockingjay}
A.~T. {Liu}, S.~{Yang}, P.~{Chi}, P.~{Hsu}, and H.~{Lee}, ``Mockingjay:
  Unsupervised speech representation learning with deep bidirectional
  transformer encoders,'' in \emph{ICASSP}, 2020, pp. 6419--6423.

\bibitem{Lian2019}
Z.~Lian, J.~Tao, B.~Liu, and J.~Huang, ``Unsupervised representation learning
  with future observation prediction for speech emotion recognition,'' in
  \emph{Interspeech}, 2019, pp. 3840--3844.

\bibitem{ling2019deep}
S.~{Ling}, Y.~{Liu}, J.~{Salazar}, and K.~{Kirchhoff}, ``Deep contextualized
  acoustic representations for semi-supervised speech recognition,'' in
  \emph{ICASSP}, 2020, pp. 6429--6433.

\bibitem{wang2020unsupervised}
W.~{Wang}, Q.~{Tang}, and K.~{Livescu}, ``Unsupervised pre-training of
  bidirectional speech encoders via masked reconstruction,'' in \emph{ICASSP},
  2020, pp. 6889--6893.

\bibitem{8682586}
Y.~{zhao}, J.~{Li}, X.~{Wang}, and Y.~{Li}, ``The speechtransformer for
  large-scale mandarin chinese speech recognition,'' in \emph{ICASSP}, 2019,
  pp. 7095--7099.

\bibitem{Ren2015FasterRT}
S.~Ren, K.~He, R.~B. Girshick, and J.~Sun, ``Faster r-cnn: Towards real-time
  object detection with region proposal networks,'' \emph{IEEE Transactions on
  Pattern Analysis and Machine Intelligence}, vol.~39, pp. 1137--1149, 2015.

\bibitem{paszke2019pytorch}
A.~Paszke, S.~Gross, F.~Massa, A.~Lerer, J.~Bradbury, G.~Chanan, T.~Killeen,
  Z.~Lin, N.~Gimelshein, L.~Antiga \emph{et~al.}, ``Pytorch: An imperative
  style, high-performance deep learning library,'' in \emph{NeurIPS}, 2019, pp.
  8024--8035.

\bibitem{povey2011kaldi}
D.~Povey, A.~Ghoshal, G.~Boulianne, L.~Burget, O.~Glembek, N.~Goel,
  M.~Hannemann, P.~Motlicek, Y.~Qian, P.~Schwarz \emph{et~al.}, ``The kaldi
  speech recognition toolkit,'' in \emph{ASRU}, 2011.

\bibitem{DBLP:journals/corr/KingmaB14}
D.~P. Kingma and J.~Ba, ``Adam: {A} method for stochastic optimization,'' in
  \emph{ICLR}, 2015.

\bibitem{Panayotov2015LibrispeechAA}
V.~Panayotov, G.~Chen, D.~Povey, and S.~Khudanpur, ``Librispeech: An asr corpus
  based on public domain audio books,'' in \emph{ICASSP}, 2015, pp. 5206--5210.

\bibitem{Rousseau2012TEDLIUMAA}
A.~Rousseau, P.~Del{\'e}glise, and Y.~Est{\`e}ve, ``Ted-lium: An automatic
  speech recognition dedicated corpus,'' in \emph{LREC}, 2012, pp. 125--129.

\bibitem{Michalek2018ASO}
J.~Michalek and J.~Van{\v{e}}k, ``A survey of recent dnn architectures on the
  timit phone recognition task,'' in \emph{ICTSD}, 2018, pp. 436--444.

\bibitem{Hinton2012DeepNN}
G.~E. Hinton, L.~Deng, D.~Yu, G.~E. Dahl, A.~rahman Mohamed, N.~Jaitly, A.~W.
  Senior, V.~Vanhoucke, P.~Nguyen, T.~N. Sainath, and B.~Kingsbury, ``Deep
  neural networks for acoustic modeling in speech recognition: The shared views
  of four research groups,'' \emph{IEEE Signal Process. Mag.}, vol.~29, pp.
  82--97, 2012.

\bibitem{li2018visualizing}
H.~Li, Z.~Xu, G.~Taylor, C.~Studer, and T.~Goldstein, ``Visualizing the loss
  landscape of neural nets,'' in \emph{NeurIPS}, 2018, pp. 6389--6399.

\bibitem{hao2019visualizing}
Y.~Hao, L.~Dong, F.~Wei, and K.~Xu, ``Visualizing and understanding the
  effectiveness of {BERT},'' in \emph{EMNLP-IJCNLP}, 2019, pp. 4143--4152.

\bibitem{Shan2019InvestigatingES}
C.~Shan, C.~Weng, G.~Wang, D.~Su, M.~X. Luo, D.~Yu, and L.~Xie, ``Investigating
  end-to-end speech recognition for mandarin-english code-switching,'' in
  \emph{ICASSP}, 2019, pp. 6056--6060.

\end{thebibliography}


\end{document}